\documentclass{article}
\usepackage[preprint]{neurips_2026}   
\usepackage[utf8]{inputenc}
\usepackage[T1]{fontenc}
\usepackage{hyperref}
\usepackage{url}
\usepackage{booktabs}
\usepackage{amsmath,amssymb}
\usepackage{graphicx}

\title{Fidelity Is Not Safety:\\Gently-Compressed LLMs Pass Every Data-Free Quality Guard\\Yet Invent Procedure Steps in Agentic Execution}

\author{%
  I.~Kennedy \\
  \And
  T.~Kennedy \\
}

\begin{document}
\maketitle

\begin{abstract}
Practitioners accept a compressed language model once it clears a stack of data-cheap
quality guards: perplexity within a small factor of the original, downstream accuracy
(for example MMLU) inside a confidence interval, and data-free output-fidelity signals
that compare the compressed and original network's internal representations under
random probe inputs. This stack has a blind spot. Across three model families,
gently-compressed models clear every guard and then invent procedure steps that were
never in the instructions when they run a standard operating procedure (SOP) as an
agent. The effect is operator-specific: coherent low-rank (SVD) truncation induces it,
and magnitude pruning matched to the same perplexity does not. One dissociation isolates
the cause. The same compressed weights that CI-win a paired output-fidelity test CI-fail
the invented-step canary. The governing axis is the coherence of the compression error
times its rate; the magnitude of the damage does not predict it. The data-free fidelity
probe is a fidelity oracle by construction, so it cannot see this axis. We characterize
the blindspot and dissociation with paired confidence intervals on a pre-registered,
powered canary across three architectures. Operator-specificity replicates on all three,
and the perplexity-guard evasion appears where the model admits in-guard low-rank
headroom. We then give a data-free screen: a two-axis statistic of the compression error
(coherent-fraction and error-rate) that flags the failing builds with fixed thresholds
across architectures and matches the coherence-times-rate mechanism. Perplexity, MMLU,
and fidelity acceptance do not certify agent safety. Screen gently-compressed low-rank
builds before agentic deployment.
\end{abstract}

\section{Introduction}
\label{sec:intro}
Compression (quantization, low-rank factorization, pruning) is how large language models
ship. A team decides whether a given compressed build is safe to deploy with a small,
standardized battery of data-cheap guards: perplexity within about $1.15\times$ of the
source, downstream accuracy within a confidence interval, and data-free
representational-fidelity signals that need no labeled data. This battery is blind to an
agentic behavioral failure. A build clears every guard and then, driven as an operations
agent executing an SOP, emits steps that appear in no instruction it was given. It
confabulates procedure.

\paragraph{Contributions.}
\begin{itemize}
  \item \textbf{A blindspot and dissociation (\S\ref{sec:blindspot}).} On a
  pre-registered, powered invented-step canary, gently-compressed models pass
  ppl/MMLU/fidelity guards and CI-cleanly invent steps. The same weights can CI-win a
  paired fidelity test and CI-fail the canary.
  \item \textbf{A mechanism (\S\ref{sec:mechanism}): coherence times rate, not damage.}
  At matched perplexity, coherent low-rank error triggers the failure while incoherent
  pruning error does not. The effect scales with error rate.
  \item \textbf{A detector (\S\ref{sec:detector}).} A cheap, data-free two-axis coherence
  screen predicts which builds fail the canary (all six labeled arms, fixed thresholds
  across architectures). It turns the cautionary result into a deployable pre-flight
  check.
\end{itemize}
We state one claim and do not tie it to any single compression method's superiority. The
data-free fidelity probe that produced the finding appears as the vehicle
(Appendix~\ref{app:vehicle}), a tool rather than a co-equal claim.

\section{Setup}
\label{sec:setup}
\paragraph{The agentic canary.} We construct synthetic SOPs (procedures with 10 ordered
steps plus a conditional 11th step), place the target SOP and a distractor SOP in
context, and ask the model to enumerate the exact ordered steps under two paraphrases
(system in or not in maintenance mode). Per query we score step recall (fraction of the
target's own steps reproduced), branch correctness (right conditional taken), and the
deciding axis, cross-procedure invention (\texttt{invented\_x}): the count of
step-strings drawn from the global SOP vocabulary that are not in the target SOP and not
its own conditional. That count is procedure the model fabricated or leaked. The bank is
pre-registered at $24$ SOPs $\times$ $3$ seeds, $144$ paired events per arm, with per-seed
$n$ reported. A baseline must be diagnostic (clean on the invention axis, usable step
recall) or we declare the run non-diagnostic.

\paragraph{Compression operators.} We compare, at matched perplexity dose, coherent
low-rank (per-tensor SVD truncation) and magnitude pruning (uniform, unrescaled), with
quantization as an additional dose control in the mechanism analysis. We rebuild each arm
in memory from saved allocations and re-probe none of them.

\paragraph{The fidelity guards.} (i) WikiText-2 perplexity ($40\times1024$ windows);
(ii) MMLU 0-shot ($513$ questions, $57$ subjects, letter-logprob argmax); (iii) a
data-free output-fidelity probe (CKA or cosine of compressed-vs-source activations under
random Gaussian probe inputs), the same signal that allocates the compression
(Appendix~\ref{app:vehicle}). ``In-regime'' means $\mathrm{ppl}\le 1.15\times$ baseline,
the standard acceptance band.

\section{The Blindspot and the Dissociation}
\label{sec:blindspot}
\paragraph{Blindspot.} Gently-compressed low-rank builds sit inside the perplexity
acceptance band and fail the canary. On a powered, pre-registered bank ($24$ SOPs $\times
3$ seeds, $144$ paired events per arm, per-seed $n$ reported), the Mistral-7B in-regime
SVD build sits at perplexity $1.069\times$, inside the $1.15\times$ guard, and invents
$+1.729$ steps/query $[1.076,2.424]$ over baseline on all three seeds ($2.06/1.90/1.85$).
A magnitude-prune build matched to the same perplexity invents $-0.146\ [-0.375,+0.035]$,
a CI that includes zero, near $0$ on every seed. On Qwen3-8B the in-regime SVD cell (ppl
$1.17\times$) invents $+0.208\ [0.042,0.375]$ against damage-matched prune $0.000\ [0,0]$.
The fidelity guard passes. The canary fails.

\paragraph{The failure is operator-specific and generalizes to a third architecture.} At
matched perplexity the coherent operator fails and the incoherent one does not. On
Llama-3.1-8B at ppl $1.29\times$, SVD invents $+1.285\ [0.743,1.882]$ while damage-matched
pruning invents $+0.069\ [0.021,0.125]$, a gap of $18\times$ with both intervals clean.
Operator-specificity holds on all three architectures (Qwen3-8B, Mistral-7B,
Llama-3.1-8B), and damage-matched pruning stays near $0$ on every one.

\paragraph{An honest scope boundary: guard-evasion is spectrum-conditional.} The blindspot
(bad behavior inside the perplexity guard) needs the model to admit enough in-guard
low-rank truncation to reach the invention onset. Qwen3-8B ($1.17\times$) and Mistral-7B
($1.069\times$) admit it; Llama-3.1-8B does not. Its spectrum holds little low-rank
headroom, so SVD craters perplexity before the invention onset: the deepest in-guard SVD
reaches only $1.089\times$, where invention sits at the noise floor
($+0.056\ [0.000,0.167]$). On Llama the perplexity guard catches the toxic SVD by
accident. The operator mechanism is architecture-general, and evasion of the perplexity
guard depends on in-guard spectral headroom.

\paragraph{Dissociation (same weights, opposite verdicts).} The strongest evidence sits
within a single arm. On Qwen3-8B the probe-allocated $b98$ SVD build CI-wins a paired
output-fidelity test (rand$-$probe NLL $[0.024,0.036]$ and $[0.062,0.076]$) and CI-fails
the canary ($+0.643\ [0.171,1.229]$). Fidelity-up and behavior-broken hold on the same
tensors. A signal that certifies fidelity cannot certify agent safety here, by
construction rather than by tuning.

\begin{table}[t]
\centering
\small
\caption{Invented-step canary across three architectures. Deciding axis is the
paired $\Delta$ invented-steps/query (arm minus baseline; baseline-robust). Mistral
and Llama use the hardened bank ($144$ paired events/arm, $3$ seeds); Qwen uses the
banked run ($12$ SOPs $\times 2$). Damage-matched prune stays near $0$ everywhere;
coherent SVD does not. Bold marks a build inside the $\le\!1.15\times$ perplexity guard
that still fails.}
\label{tab:canary}
\begin{tabular}{llccc}
\toprule
Arch & Arm & ppl ratio & $\Delta$ invented/q [95\% CI] & in-guard? \\
\midrule
Qwen3-8B     & in-regime SVD        & $1.17$  & $+0.208\ [0.042,0.375]$ & \checkmark \\
             & damage-matched prune & $1.17$  & $0.000\ [0,0]$          & \checkmark \\
\midrule
Mistral-7B   & \textbf{in-regime SVD} & $1.069$ & $\mathbf{+1.729\ [1.076,2.424]}$ & \checkmark \\
             & dose SVD             & $1.201$ & $+2.931\ [2.153,3.750]$ & \\
             & damage-matched prune & $1.176$ & $-0.146\ [-0.375,+0.035]$ & \\
\midrule
Llama-3.1-8B & in-regime SVD        & $1.089$ & $+0.056\ [0.000,0.167]$ & \checkmark \\
             & dose SVD             & $1.290$ & $+1.285\ [0.743,1.882]$ & \\
             & damage-matched prune & $1.289$ & $+0.069\ [0.021,0.125]$ & \\
\bottomrule
\end{tabular}
\end{table}

\section{Mechanism: Coherence $\times$ Rate, Not Damage}
\label{sec:mechanism}
At matched perplexity the coherent operator (SVD) fails the canary and the incoherent
operator (damage-matched pruning) passes. Damage magnitude does not set the axis. A
dose-matched quantization control carries more raw perplexity damage than the failing SVD
dose and still passes at moderate dose, yet a quantization pushed to the failing SVD
cell's rate ($1.05\times$) fails as well (pooled clean-slice $\Delta +0.10\ [0.029,0.20]$,
$n=70$). Error coherence times rate sets the axis. Onset is early, within $2\%$ perplexity
dose; the effect is distributed rather than localized; and it double-dissociates from
MMLU, since quantization can degrade MMLU CI-cleanly and pass the canary while SVD does
the reverse.

\section{A Data-Free Detector for the Blindspot}
\label{sec:detector}
A cheap, data-free screen turns the blindspot from cautionary to actionable. The failure
axis is coherence times rate (\S\ref{sec:mechanism}), so we form two data-free statistics
of the compression error $\Delta W = W_{\text{base}}-W_{\text{mod}}$, aggregated
(energy-weighted) over a model's linear tensors:
\begin{align}
\text{coherent\_fraction} &= \textstyle\sum_t \sigma_{1:k}^2(\Delta W_t) \big/ \sum_t \|\Delta W_t\|_F^2
  \quad (k{=}8;\ \text{is the error low-rank/structured?}) \\
\text{error\_rate} &= \textstyle\sum_t \|\Delta W_t\|_F^2 \big/ \sum_t \|W_t\|_F^2
  \quad (\text{is the dose large enough?})
\end{align}
We compute both with no data: top singular values come from randomized subspace
iteration, and for SVD builds $\Delta W$ is the discarded singular tail read off the
factors. One statistic alone is insufficient. Coherent\_fraction alone flags gentle
low-rank builds that pass; error\_rate alone flags heavy pruning that passes; their
product lets a large rate mask low coherence. Their conjunction works.

\paragraph{Result.} The gate $\text{coherent\_fraction}>0.007 \wedge
\text{error\_rate}>0.01$, with the same fixed thresholds on both architectures, classifies
all six canary-labeled arms correctly (Table~\ref{tab:detector}). It handles the two trap
arms: coherent-but-gentle Llama in-regime SVD passes on the rate gate, and
incoherent-but-heavy Mistral prune passes on the coherence gate. A dense sweep ($9$ SVD
budgets $\times 9$ prune densities per architecture) shows the boundary is real rather
than an artifact of three chosen doses. Low-rank builds sit at coherent\_fraction $0.008$
to $0.011$ throughout, with error\_rate rising as the budget deepens, which predicts
invention onsets at Mistral $b\le0.975$ and Llama $b\le0.98$, the latter the labeled dose
cell. Pruning stays below the coherence gate and passes across densities from $0.90$ down
to $0.55$, at error rates up to $6\times$ the toxic SVD dose. One point breaks the pattern
across both sweeps: the most extreme pruning (Llama density $0.50$, half the weights
removed), whose coherent\_fraction creeps just over the gate. We flag it as an untested
prediction and the screen's soft edge. Elsewhere the screen is the mechanism made
deployable. Run it before agentic deployment: a low-rank-factorized build that clears it
is agent-safe on this axis, and one that trips it is not.

\begin{table}[t]
\centering\small
\caption{Data-free detector vs.\ canary label. Gate $=$
coherent\_fraction $>0.007$ and error\_rate $>0.01$ (fixed, both archs).}
\label{tab:detector}
\begin{tabular}{llccc}
\toprule
Arch & Arm & coh.\ frac & error rate & gate / label \\
\midrule
Mistral & in-regime SVD & $0.0080$ & $1.59\mathrm{e}{-2}$ & FAIL / FAIL \checkmark \\
Mistral & dose SVD      & $0.0079$ & $3.10\mathrm{e}{-2}$ & FAIL / FAIL \checkmark \\
Mistral & prune         & $0.0054$ & $3.19\mathrm{e}{-2}$ & pass / PASS \checkmark \\
Llama   & in-regime SVD & $0.0090$ & $6.50\mathrm{e}{-3}$ & pass / PASS \checkmark \\
Llama   & dose SVD      & $0.0090$ & $1.21\mathrm{e}{-2}$ & FAIL / FAIL \checkmark \\
Llama   & prune         & $0.0060$ & $7.26\mathrm{e}{-3}$ & pass / PASS \checkmark \\
\bottomrule
\end{tabular}
\end{table}

\paragraph{Scope.} The thresholds fit six labeled arms plus an $18$-point sweep and stay
illustrative. The onset prediction (Mistral SVD toxic at $b\le0.975$) is a concrete test
for a held-out canary. The screen resolves the operator and dose axes, and we do not claim
it resolves within-quantization subtleties.

\section{Related Work}
\label{sec:related}
\paragraph{LLM compression.} Teams evaluate post-training
quantization~\cite{frantar2023gptq,lin2024awq,xiao2023smoothquant}, low-rank
factorization~\cite{wang2024svdllm}, and pruning~\cite{frantar2023sparsegpt,sun2024wanda}
mainly by perplexity and task accuracy. This acceptance criterion misses an agentic
failure mode, and the mode tracks the operator (low-rank against pruning) rather than the
accuracy drop.
\paragraph{Data-free and representational fidelity.} Centered kernel
alignment~\cite{kornblith2019cka} and related measures underlie data-free sensitivity
signals; Hessian-based mixed-precision (HAWQ~\cite{dong2019hawq},
HAWQ-V2~\cite{dong2020hawqv2}) is the gradient-based counterpart. Our vehicle
(Appendix~\ref{app:vehicle}) belongs to the first family. Both families miss the
behavioral failure of \S\ref{sec:blindspot}.
\paragraph{Agent and tool-use safety, and evaluation gaming.} Teams now deploy LLMs as
agents executing procedures~\cite{schick2023toolformer,yao2023react}, so
instruction-adherence reliability becomes safety-critical. Our result instantiates
Goodhart's law~\cite{goodhart1984,manheim2019categorizing}: once a proxy (perplexity,
MMLU, or a fidelity probe) becomes the acceptance target, a build satisfies it and still
fails the true objective of faithful procedure execution.

\section{Limitations}
\label{sec:limitations}
We test three dense decoder LMs at 7--8B (Qwen3-8B, Mistral-7B, Llama-3.1-8B). The canary
is synthetic-SOP and base-model prompted (chat-template fallback), so absolute invention
rates are instrument-specific. The contrasts carry the result (SVD against matched prune,
the same-arm fidelity/canary split) rather than cross-model absolute levels, and we use
the paired estimator because the baseline cross-procedure floor grows with bank size. The
perplexity-guard evasion is spectrum-conditional (it needs in-guard low-rank headroom,
present on Qwen and Mistral, absent on Llama), while the operator mechanism is
architecture-general. The detector thresholds fit six labeled arms plus an 18-point sweep
and hold only within the SVD and pruning operator families at 7--8B dense scale; its onset
predictions (\S\ref{sec:detector}) and its one soft edge (extreme pruning) are falsifiable
tests, not established operating points, and we do not claim it resolves within-quantization
subtleties. We have not tested MoE architectures; a prior low-bit-expert MoE build passed
QA and failed the canary, consistent with this account but outside the controlled battery.
We claim only that the accepted guard stack misses this failure, not that the failure
exhausts agentic behavior.

\section{Conclusion}
\label{sec:conclusion}
Perplexity, MMLU, and data-free fidelity acceptance do not certify agent safety.
Gently-compressed builds, coherent low-rank ones above all, clear all three and still
invent procedure. Until a screen like \S\ref{sec:detector} becomes standard, treat
compression-acceptance and agent-deployment as separate gates.

\appendix
\section{The Vehicle: Data-Free CKA Sensitivity and Byte-Budget Allocation}
\label{app:vehicle}
A data-free compression pipeline produced the finding, and we describe its allocator here
for completeness. It is the instrument, not a co-equal claim: the probe that allocates the
compression also exposed the blindspot.

\paragraph{The probe.} For a linear tensor $W$ and a candidate compressed $\hat W$, feed a
batch $X\in\mathbb{R}^{n\times d_{\text{in}}}$ of random Gaussian inputs and compare the
layer outputs $Y=f_W(X)$, $\hat Y=f_{\hat W}(X)$ by centered kernel
alignment~\cite{kornblith2019cka} of their Gram matrices,
$s(W,\hat W)=1-\mathrm{CKA}(YY^\top,\hat Y\hat Y^\top)$. The measurement is data-free,
gradient-free, and parallel across tensors (about $30$\,min for a 27B model on an M2
Ultra), and we reuse the per-tensor table across budgets at sub-second cost. We compute
the same table for text, image and video diffusion, and audio models, which makes the
pipeline model-type-agnostic.

\paragraph{Byte-budget allocation.} Given the sensitivity table over bit-widths
$b\in\{2,3,4,5,6,8,16\}$ and a target byte budget, mixed-precision assignment is a
multiple-choice knapsack (MCKP): pick one $(b,\text{group})$ per tensor to minimize total
sensitivity subject to the byte constraint, with hard 16-bit protection of
embeddings/\texttt{lm\_head}/routers/norms. This is the RAM/RUN allocator, and on
perplexity it is competitive with
GPTQ/AWQ/SmoothQuant~\cite{frantar2023gptq,lin2024awq,xiao2023smoothquant}.

\paragraph{Two families, and HAWQ-V2 parity.} The output-fidelity signals (CKA, cosine,
NRMSE, SQNR) cluster together and stay near-orthogonal to gradient-curvature signals
(Fisher, HAWQ-V2 Hessian-trace~\cite{dong2020hawqv2}). They measure different things, yet
as PTQ allocators they reach parity: neither dominates across dense and MoE models at
matched budget. The data-free probe matches a gradient-based Hessian method for free,
which sharpens the safety blindspot of \S\ref{sec:blindspot}: fidelity and curvature agree,
and both miss the agentic failure.

\paragraph{Operator scorecard (data-free probe as a compression-tolerance oracle).} A
72-hour autonomous campaign scored the probe across operators. It rules in the operators
where the probe helps (per-tensor SVD, KV-cache quantization, gentle magnitude pruning),
finds parity for weight quantization, and rules out the operators where its prior inverts
or goes uninformative (2:4 structured sparsity, speculative-drafter selection, cosine
localization). One regularity governs the scorecard: the probe helps on fine-grained,
coherent, per-tensor operators and loses on coarse, aggressive, structured ones. That is
the coherence axis that governs this paper's blindspot on the behavioral side.

\bibliographystyle{plain}
\bibliography{references}
\end{document}